\documentclass[a4paper, 10pt, conference]{ieeeconf}      
\usepackage{FG2024}

\FGfinalcopy 

\IEEEoverridecommandlockouts                              
\usepackage{makecell}
\usepackage{graphicx}
\usepackage[T1]{fontenc}
\usepackage[utf8]{inputenc}
\usepackage{hyperref}       
\usepackage{url}            
\usepackage{booktabs}       
\usepackage{amsfonts}       
\usepackage{nicefrac}       
\usepackage{microtype}      
\usepackage{epstopdf}
\usepackage{xcolor}
\usepackage{amsmath}
\usepackage{multirow}
\usepackage{bm}
\usepackage{cite}
\usepackage{mathtools}
\def\FGPaperID{127} 

\title{\LARGE \bf
Multi-Scale Spatio-Temporal Graph Convolutional Network \\for Facial Expression Spotting 
}

\author{\parbox{16cm}{\centering
   {\large Yicheng Deng, Hideaki Hayashi, Hajime Nagahara}\\  
   {\normalsize
   Osaka University, Japan\\}}
   \thanks{This work was partially supported by Innovation Platform for Society 5.0 from Japan Ministry of Education, Culture, Sports, Science and Technology, and JSPS KAKENHI Grant Number JP21H03511.}
}

\begin{document}

\ifFGfinal
\thispagestyle{empty}
\pagestyle{empty}
\else
\author{Anonymous FG2024 submission\\ Paper ID \FGPaperID \\}
\pagestyle{plain}
\fi
\maketitle


\begin{abstract}

Facial expression spotting is a significant but challenging task in facial expression analysis. 
The accuracy of expression spotting is affected not only by irrelevant facial movements but also by the difficulty of perceiving subtle motions in micro-expressions. In this paper, we propose a Multi-\textit{S}cale S\textit{p}ati\textit{o}-\textit{T}emporal \textit{G}raph \textit{C}onvolutional \textit{N}etwork (SpoT-GCN) for facial expression spotting. To extract more robust motion features, we track both short- and long-term motion of facial muscles in compact sliding windows whose window length adapts to the temporal receptive field of the network. This strategy, termed the receptive field adaptive sliding window strategy, effectively magnifies the motion features while alleviating the problem of severe head movement. The subtle motion features are then converted to a facial graph representation, whose spatio-temporal graph patterns are learned by a graph convolutional network. This network learns both local and global features from multiple scales of facial graph structures using our proposed facial local graph pooling (FLGP). Furthermore, we introduce supervised contrastive learning to enhance the discriminative capability of our model for difficult-to-classify frames. The experimental results on the SAMM-LV and CAS(ME)$^\mathrm{2}$ datasets demonstrate that our method achieves state-of-the-art performance, particularly in micro-expression spotting. Ablation studies further verify the effectiveness of our proposed modules.
\end{abstract}

\section{INTRODUCTION}

Facial expressions are a typical form of nonverbal communication used to convey human emotions. When people undergo emotional changes, voluntary or involuntary facial muscle movements create various expressions. These expressions act as simple and direct social signals, allowing others to understand their emotions and facilitating human communication.

In general, facial expressions can be divided into two categories: macro-expressions and micro-expressions. Macro-expressions usually last for 0.5 to 4.0 seconds~\cite{nummenmaaekman}, and they are easy to perceive due to their occurrence on a large facial area and high intensity \cite{corneanu2016survey}. Macro-expression analysis is important in many practical applications such as sociable robots \cite{fukuda2002facial}, mental health \cite{kopper1996experience}, and virtual reality \cite{facialvr}. In contrast, micro-expressions generally last for less than 0.5 seconds \cite{yan2013fast}, and they are hard to perceive due to their locality and low intensity \cite{bhushan2015study}. Due to their involuntary nature, they are crucial in situations where people may want to suppress their emotions or attempt to deceive others, such as lie detection \cite{ekman2009telling}, medical care \cite{endres2009micro}, and national security \cite{o2009police}. Therefore, both macro- and micro-expression analysis play important roles in understanding human emotions and behaviors. 

\begin{figure}[t]
\centering
\includegraphics[width=0.48\textwidth]{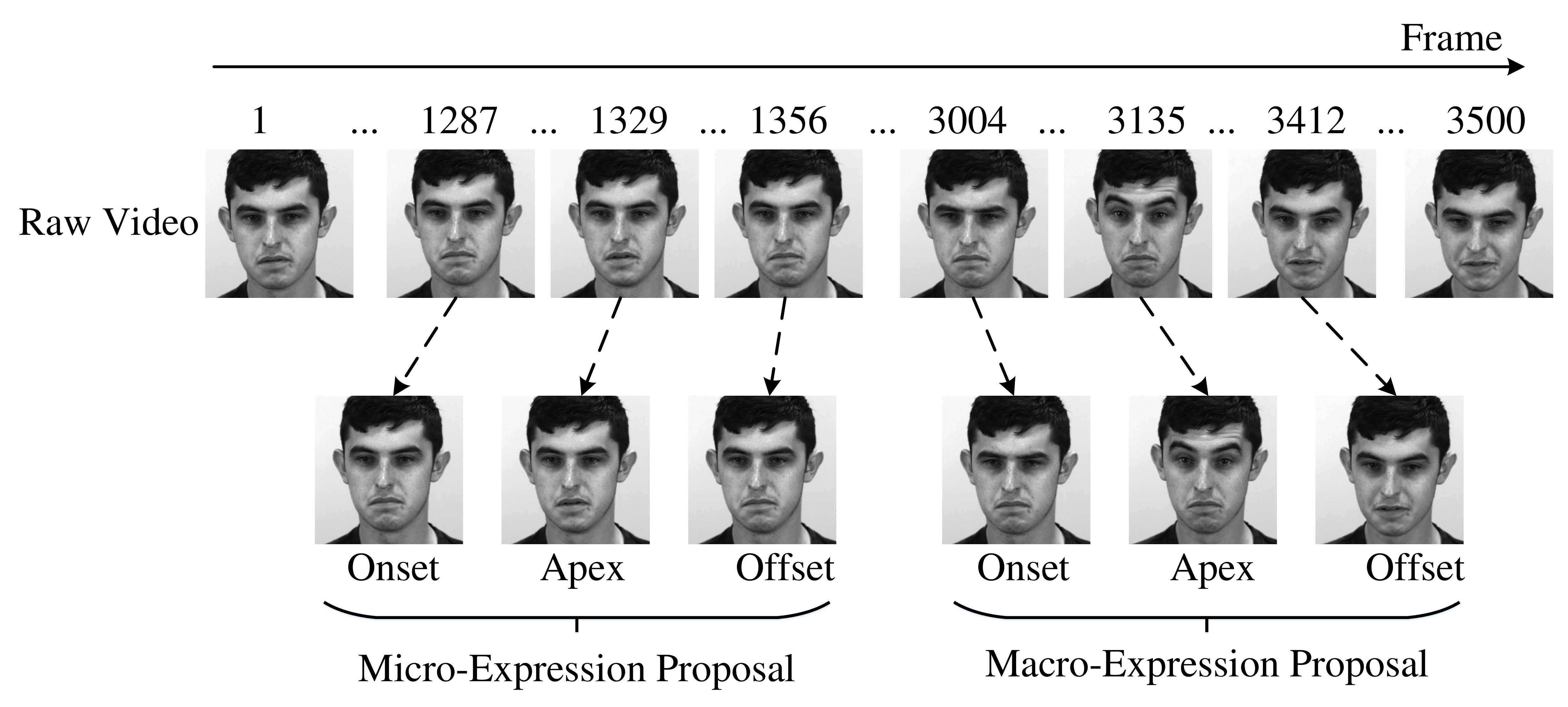}
\caption{Illustration of macro- and micro-expression spotting.}
\label{abstract}
\end{figure}

Facial expression spotting is the preliminary step in facial expression analysis, aiming to locate the onset and offset frames of macro- and micro-expression intervals in long videos, as shown in Fig.~\ref{abstract}. The onset frame represents when an expression starts, and the offset frame represents when it ends. However, the difficulty in perceiving micro-expressions and the presence of irrelevant motions in long videos, such as head movements and eye blinking, make this task highly challenging.

In recent years, many researchers have devoted a lot of effort to developing effective algorithms for facial expression spotting. Early works employed traditional methods to extract hand-crafted features, analyze feature differences, and spot expression clips using threshold strategies \cite{gan2020optical, he2020spotting, yuhong2021research, zhao2022rethinking}. 
Lately, with the development of deep learning, several learning-based methods have been proposed for expression spotting \cite{verburg2019micro, wang2021mesnet, yap20223d, yu2021lssnet, yin2023aware}, but the spotting accuracy, especially in micro-expression spotting, still needs improvement. 

The main problems arise from their choice of feature extraction strategy and the oversimplification of the network.
First, some methods utilize a large sliding window strategy to spot potential expression proposals\cite{he2020spotting, guo2021magnitude}, where the accuracy is severely affected by head movement. Alternatively, some other methods calculate optical flows between adjacent frames as motion features\cite{yu2021lssnet, leng2022abpn}, but they cannot effectively reveal subtle motions that exist in micro-expressions. Therefore, there is an urgent need to find a strategy that can magnify motion information while alleviating the influence of head movement. 
Second, some methods compute optical flows in specific regions of interest (ROIs) related to the action unit to alleviate the influence of irrelevant motions and then utilize a network to learn the extracted motion features\cite{leng2022abpn,yin2023aware}. However, their proposed network is overly simplistic, lacking a comprehensive consideration of spatio-temporal relationships and multi-scale feature learning, which limits the representational capacity of their models.

To address these issues, we propose a multi-scale spatio-temporal graph convolutional network, termed \textit{SpoT-GCN}, for macro- and micro-expression spotting. Specifically, we design a receptive field adaptive sliding window strategy, where the temporal window size corresponds to the receptive field of the network, to compute and combine short- and long-term optical flows as input for frame-level apex or boundary (onset or offset) probability estimation, amplifying the motion features while avoiding significant head movement problems. Then, we adopt a graph convolutional network to capture the spatial relationships and temporal variations among different facial parts across frames, where a facial local graph pooling (FLGP) strategy is proposed to extract multi-scale facial graph-structured features, enhancing the model's understanding ability from local to global. 

We also notice that distinguishing between certain macro-expressions and micro-expressions near the boundary is difficult. Additionally, some normal frames might be misclassified as micro-expression frames due to the noises that exist in the extracted optical flows. To address this issue, we introduce supervised contrastive learning into our model to learn finer discriminative feature representation for better distinguishing different types of frames in long videos. 

Our main contributions are as follows:
\begin{itemize}
\item We propose a novel graph convolutional network (GCN) to comprehensively capture the spatial relationships and temporal variations among different facial parts across frames, in which a graph pooling strategy suitable for facial structure is proposed for multi-scale feature learning.
\item We design a receptive field adaptive sliding window strategy to compute short- and long-term optical flows for frame-level apex or boundary probability estimation, which not only magnifies the motion information but also avoids large head movement problems.
\item We introduce supervised contrastive loss to our model for discriminative feature representation learning. To the best of our knowledge, our work is the first to study contrastive learning for facial expression spotting, achieving the recognition of boundaries between different types of expressions.
\end{itemize}

\begin{figure}[tbp]
\centering
\includegraphics[width=0.25\textwidth]{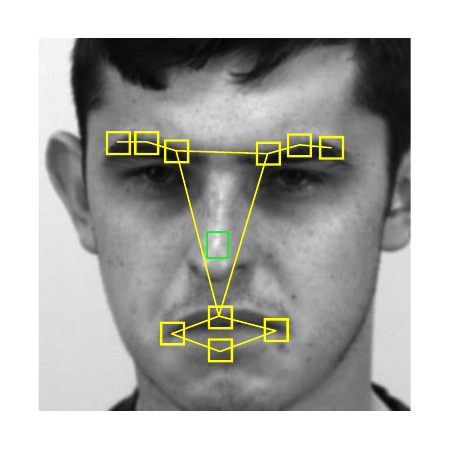}
\caption{Extracted ROIs and constructed facial graph structure are denoted in yellow, while the nose tip region for face alignment is denoted in green.}
\label{rois}
\end{figure}

\begin{figure*}[tbp]
\centering
\includegraphics[width=1.0\textwidth]{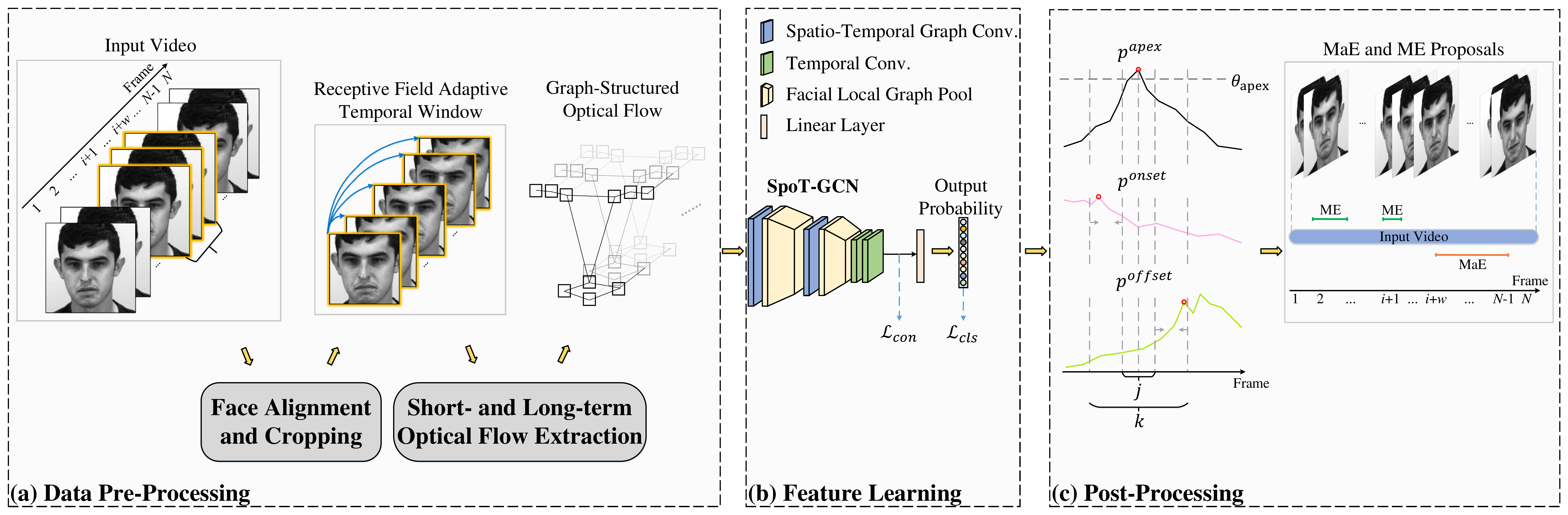}
\caption{Overview of the proposed framework. (a) The data pre-processing module partitions the input video into overlapping temporal windows using the receptive field adaptive sliding window strategy and extracts facial graph-structured optical flows; (b) the feature learning module employs the SpoT-GCN which takes optical flow features as input for frame-level apex or boundary probability estimation; (c) the post-processing module aggregates the probability maps from all frames and generates expression proposals.}
\label{framework}
\end{figure*}

\section{RELATED WORKS}

\subsection{Facial expression spotting}
Current facial expression spotting methods can be divided into two categories: traditional methods and deep learning methods.
Early works employed appearance-based feature extraction techniques, such as local binary patterns \cite{moilanen2014spotting} and histogram of oriented gradients \cite{davison2015micro}, and combined them with machine learning algorithms for feature difference analysis, then utilizing threshold strategies for expression spotting.
Subsequently, using the optical flow algorithm to extract motion features became mainstream. He et al.\cite{he2020spotting} employed the method of main directional maximal difference analysis \cite{wang2017main} to utilize the maximal difference in magnitude along the main direction of optical flow features to detect facial movements.
He~\cite{yuhong2021research} proposed computing optical flows for more accurate face alignment and performed the spotting process using a sliding window strategy. 

With the development of deep learning, several researchers proposed various neural networks for feature learning.
Zhang et al. \cite{zhang2018smeconvnet} utilized convolutional neural networks (CNNs) to extract features from video clips and spotted the apex frames from long videos using a feature matrix processing method.
Liong et al. \cite{liong2021shallow} introduced an optical flow-based three-stream CNN and a pseudo-labeling technique to facilitate the learning process.
Yu et al. \cite{yu2021lssnet} introduced a two-stream CNN network to extract features from raw videos and optical flow maps. Then they added location suppression modules to the network to reduce redundant proposals. However, the spotting accuracy may be affected by irrelevant motions when processing entire images without any masking.
Yang et al. \cite{yang2021facial} utilized facial action unit information and concatenated various types of neural networks for feature learning.
Leng et al. \cite{leng2022abpn} extracted 12 ROIs and adopted the main directional mean optical flow algorithm \cite{liu2015main} to compute optical flows between adjacent frames. Then they utilized one-dimensional CNNs to learn temporal variations and predicted the probability that each frame belongs to an apex or boundary frame. 
Based on \cite{leng2022abpn}, Yin et al. \cite{yin2023aware} learned spatial relations by adding a GCN to embed action units (AUs) label information into the extracted optical flows. While current optical flow-based methods have achieved a significant improvement in MaE spotting, the performance in ME spotting remains considerably lower. This is because their extracted optical flow features cannot reveal subtle motions that exist in micro-expressions. Our receptive field adaptive sliding window strategy can effectively magnify these subtle motions, thereby improving ME spotting performance.

\subsection{Graph Convolutional Network}
Kipf et al. \cite{kipf2016semi} proposed GCNs in 2017, presenting an effective convolutional operation that captures relationships between nodes, thus enabling deep learning on graph-structured data. In recent years, researchers have begun to apply GCNs to human emotion analysis, considering the human face as a graph. GCNs are beneficial for facial expression analysis, as the issue of irrelevant facial muscle movements can be alleviated by extracting only several ROIs instead of processing entire human face images. Liu et al. \cite{liu2020relation} were the first to utilize GCNs for action unit detection. They treated each AU-related region as a graph node and employed a GCN to learn the AU relations. Jin et al. \cite{jin2021learning} presented a double dynamic relationships GCN for facial expression recognition (FER). Liu et al. \cite{liu2023facial} proposed to recognize facial expressions by combing the advantages of CNN for extracting features and GCN for modeling complex graph patterns to capture the underlying relationship between expressions, thus improving the accuracy and efficiency of FER. Xie et al. \cite{xie2020assisted} explored the application of GCNs in micro-expression recognition and introduced a GCN for AU relation learning. Kumar et al.\cite{kumar2023relational} presented a two-stream relational edge-node graph attention network to improve the accruracy of micro-expression recognition. Yin et al. \cite{yin2023aware} were the first to apply GCNs to macro- and micro-spotting tasks. However, they simply used GCNs to learn spatial information, then they flattened the facial graph-structured data and fed it into CNNs to learn temporal information, which fails to comprehensively capture the spatio-temporal dependencies among different facial parts. To solve this problem, we propose SpoT-GCN, which considers the spatial relationships of multi-scale facial graph structures as well as the temporal variations of various facial parts at different scales, significantly enhancing the model's representational capacity.

\subsection{Contrastive learning}
In recent years, contrastive learning has demonstrated significant effectiveness in the field of unsupervised representation learning. The goal of contrastive learning is to learn meaningful representations by maximizing the similarity between positive pairs and minimizing it between negative pairs. Chen et al. \cite{chen2020simple} introduced unsupervised contrastive learning and generated positive pairs using data augmentation. This approach yields meaningful visual representations that can subsequently be applied in downstream tasks like image recognition and image segmentation.
Then, Khosla et al. \cite{khosla2020supervised} proposed the supervised contrastive (SupCon) loss, showcasing its potential to enhance supervised tasks by introducing class information into the contrastive loss. Supervised contrastive learning efficiently enlarges the domain discrepancy, thereby improving discriminative feature representation extraction. Our method aims to establish the contrast between macro- and micro-expression frames, as well as between micro-expression frames and normal frames. This approach allows us to learn more discriminative feature representations and consequently reduce the misclassification rate.



\section{METHODOLOGY}
Given a long video as input, our goal is to spot all potential macro- and micro-expression intervals within the video, locating the onset and offset frames as well as determining the expression type for each expression proposal.
As illustrated in Fig.~\ref{framework}, our framework comprises three modules: data pre-processing module, feature learning module, and post-processing module.

\subsection{Data pre-processing}

\begin{figure*}[tbp]
\centering
\includegraphics[width=1.0\textwidth]{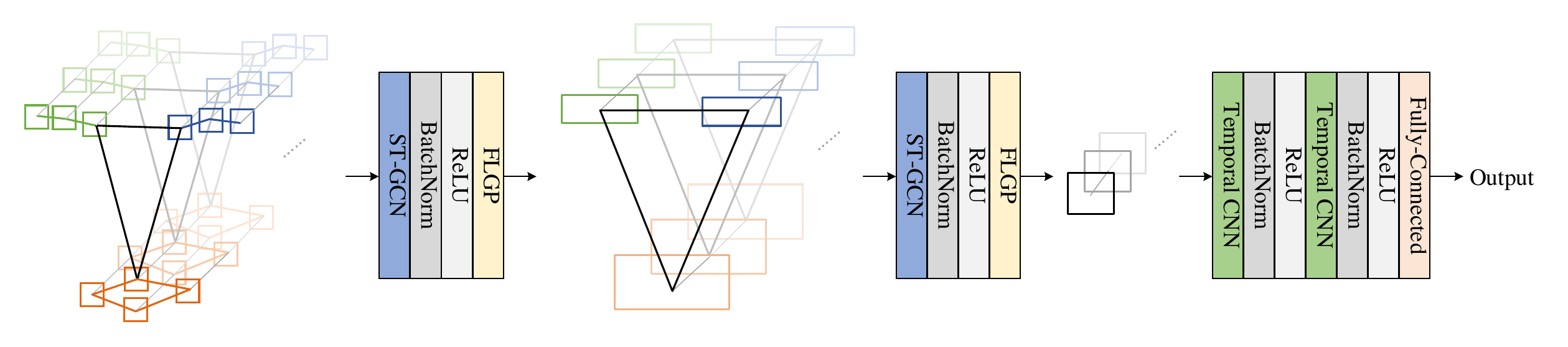}
\caption{Network structure of our SpoT-GCN and the scale change between different facial graph structures through FLGP.}
\label{netandflgp}
\end{figure*}

Suppose we have a raw video $V=(v_i)_{i=1}^{N}$ as input, where $N$ represents the total number of frames in the video. 
First, given a window length $w$, we pad the beginning of the video with $\lfloor \frac{w}{2} \rfloor$ repetitions of the first frame and pad the end of the video with $\lfloor \frac{w}{2} \rfloor$ repetitions of the last frame. Then, different from previous methods that treat the receptive field and temporal sliding window size separately, we employ a receptive field adaptive sliding window strategy with a window length $w$ and a stride of $1$ to partition the video into multiple overlapping clips $C=(c_i)_{i=1}^N$, where $w$ adapts to the temporal receptive field of the network and $c_i$ corresponds to the clip whose all temporal information will be fully utilized for predicting the $i$-th frame $v_i$. 

For each clip $c_i$, we employ MobileFaceNet~\cite{PFL} to detect 68 facial key points in the first frame $c_i^1$. These key points are then used for cropping the human face, extracting the nose region for face alignment, and extracting ROIs. Specifically, we detect the facial bounding box in the first frame. Then, for each subsequent frame $c_i^s, s=2, 3, \ldots, w$, we initialize the facial bounding box using that of the first frame. Afterward, since the nose tip area remains stationary during expressions, we compute the optical flow of the nose tip region, shown in Fig.~\ref{rois}, as the global head movement and use it to adjust the facial bounding box for $c_i^s$. Specifically, we
utilize the Farneback algorithm \cite{farneback2003two} to compute the average optical flow of the nose tip region $o_i^{s,\mathrm{nose}}\in\mathbb{R}^{2}$ between $c_i^1$ and $c_i^s$:
\begin{equation}
    o^{s,\mathrm{nose}}_i=\frac{1}{m^{\mathrm{nose}}_i\times n^{\mathrm{nose}}_i}\sum_{(x,y)\in M^{\mathrm{nose}}_i}\mathrm{OF}(c_i^1, c_i^s),
\label{of}
\end{equation}
where $m^{\mathrm{nose}}_i$ and $n^{\mathrm{nose}}_i$ represent the height and width of the nose tip region $M^{\mathrm{nose}}_i$, and $\mathrm{OF}(\cdot, \cdot)$ represents the computation of optical flow. 

Since processing entire images could affect prediction accuracy due to irrelevant motion (e.g., eye blink) and background information, we selectively extract $R=10$ ROIs that are most representative of facial expressions. The extracted ROIs are shown in Fig.~\ref{rois}. We compute the optical flow for the chosen $R$ ROIs between $c_i^1$ and $c_i^s$ to obtain optical flow features $o_i^s=[o_i^{s,r}]_{r=1}^R\in\mathbb{R}^{R\times 2}$.
For each ROI, the optical flow computation is similar to (\ref{of}). Then we construct $o_i\in\mathbb{R}^{w\times R\times 2}$ for the $i$-th clip $c_i$ by concatenating the optical flows $[o_i^1, o_i^2, \ldots, o_i^w]$, where $o_i^1=\bm{0}$.
As a result of the data pre-processing, we obtain the optical flow features $O=(o_i)_{i=1}^N\in\mathbb{R}^{N\times w\times R\times 2}$ for the entire input video.


\subsection{Multi-scale spatio-temporal GCN for feature learning}
After obtaining optical flow features, we employ the proposed SpoT-GCN for feature learning. The network structure is shown in Fig. \ref{netandflgp}. We first use spatio-temporal GCNs (ST-GCNs) to capture spatio-temporal relationships among different ROIs across temporal frames.
Just as standard three-dimensional (3D) CNNs can be seen as learning kernels to discover meaningful and distinguishable latent patterns in videos, ST-GCNs acquire knowledge about embedded nodes based on spatio-temporal neighbors and graph relations. In our method, an ST-GCN takes the graph node optical flow features $O$ and the adjacency matrix $A$ as input.
To this end, we build the initial adjacency matrix $A\in\mathbb{R}^{R\times R}$ empirically based on the designed facial graph structure shown in Fig.~\ref{rois}, where the weights of all edges are set to 1, as we only use the edges to represent the facial spatial structure. In practice, we stack $G$ ST-GCN layers. For the $g$-th GCN layer, its operation can be expressed as:
\begin{equation}
    H_g=\sigma(AH_{g-1}W_{g}),
\end{equation}
where $\sigma$ is the activation function, $H_{g-1}$ is the output of the $(g-1)$-th ST-GCN layer, and $W_{g}$ are the learnable weights of the $g$-th ST-GCN layer.


Multi-scale learning is significant and has shown powerful performance in CNN-based image processing tasks \cite{lin2017feature, liu2021swin} because it enables the model to extract richer and more diverse feature representations at various scales. However, applying pooling operations, which are generally used for downsampling images, to graph-structured data presents challenges because it is a type of non-Euclidean structured data. To address this issue, inspired by Xu et al. \cite{xu2021graph}, we introduce FLGP, specifically designed for extracting multi-scale facial graph features. In practice, we design three scales of facial structures. The designed scales and the scale change achieved through FLGP are illustrated in Fig.~\ref{netandflgp}. Note that we only use FLGP to downsample the spatial scale while maintaining the temporal dimension, relying on the network layers with learnable weights to fully learn the temporal dynamics. During each FLGP operation, the facial graph is downsampled by aggregating features of several nodes using max pooling after each ST-GCN layer. Finally, we aggregate the global spatial features into a single node.

Since we do not need to learn spatial relationships anymore after downsampling the scale of the facial graph to a single node, we employ temporal convolutional networks (TCNs) to capture the remaining high-level temporal variations. In practice, we stack $C$ TCN layers and the output of the $c$-th layer can be expressed as:
\begin{equation}
    H_c=\sigma(H_{c-1}W_{c}),
\end{equation}
where $\sigma$ is the activation function, $H_{c-1}$ is the output of the $(c-1)$-th TCN layer, and $W_{c}$ are the learnable weights of the $c$-th TCN layer. 

Finally, the single remaining node fully aggregates spatio-temporal relationships in a sliding window, which is then fed into a fully-connected layer to output the probability map $p_i=\{p_i^{\mathrm{onset}}$, $p_i^{\mathrm{apex}}$, $p_i^{\mathrm{offset}}$, $p_i^{\mathrm{exp}}$, $p_i^{\mathrm{norm}}$\} for the frame $v_i$. This map contains the probabilities of $v_i$ being an onset frame, apex frame, offset frame, expression frame, or normal frame. In addition, each component in $p_i$ includes two probabilities for micro-expression spotting and macro-expression spotting, respectively. Specifically, $p_i^{\mathrm{onset}}=\{p_i^{\mathrm{mi},\mathrm{onset}}, p_i^{\mathrm{ma},\mathrm{onset}}\}$, $p_i^{\mathrm{apex}}=\{p_i^{\mathrm{mi},\mathrm{apex}}, p_i^{\mathrm{ma},\mathrm{apex}}\}$, $p_i^{\mathrm{offset}}=\{p_i^{\mathrm{mi},\mathrm{offset}}, p_i^{\mathrm{ma},\mathrm{offset}}\}$, $p_i^{\mathrm{exp}}=\{p_i^{\mathrm{mi},\mathrm{exp}}, p_i^{\mathrm{ma},\mathrm{exp}}\}$, $p_i^{\mathrm{norm}}=\{p_i^{\mathrm{mi},\mathrm{norm}}, p_i^{\mathrm{ma},\mathrm{norm}}\}$. We split the optimization tasks into two binary classification tasks and a three-class classification task for different types of frames, following the optimization method outlined in \cite{leng2022abpn}.
We employ focal-loss \cite{lin2017focal} to optimize our model, which can be expressed as:
\begin{equation}
    \mathcal{L}_{\mathrm{cls}} = -\sum_{i} y_i\alpha (1 - p_i)^\gamma \log(p_i),
\end{equation}
where $y_i$ is the ground-truth label, $\alpha$ and $\gamma$ are hyperparameters, respectively.

\subsection{Supervised contrastive learning}
Until now, our focus has been on minimizing the divergence between the predicted class probabilities and the ground-truth class labels, which might neglect the distributional differences among different classes.
Some macro-expressions and micro-expressions near the boundary are hard to distinguish in terms of duration and intensity. For example, the labeling of macro- and micro-expressions follows a rule where macro-expressions have a duration greater than 0.5 seconds, and micro-expressions have a duration smaller than 0.5 seconds. This rule creates difficulty in distinguishing frames of expressions whose duration is close to 0.5 seconds. Similar issues also exist when distinguishing between micro-expression frames and normal frames. This is due to the fact that some ground-truth micro-expressions exhibit very low intensity, and the noises exist in optical flow features can lead to the misclassification of certain normal frames as micro-expression frames.

To address this issue, we introduce supervised contrastive learning \cite{khosla2020supervised} to enhance the discriminative feature learning for classifying different types of frames. Specifically, we utilize the output of the final TCN layer as the feature representation for each frame. Then we introduce a supervised contrastive loss to minimize the distance between feature representations of the same class while simultaneously pushing apart feature representations of different classes. We use the frame type label $\widetilde{y}_i$ for the $i$-th sample to provide supervision for the supervised contrastive loss. This means that each frame is labeled as a macro-expression frame, micro-expression frame, or normal frame. Let $I$ denote a set of samples in a batch, and the loss function can be expressed as:
\begin{equation}
\begin{aligned}
\mathcal{L}_{\mathrm{con}}=\sum_{i\in I}\frac{-1}{|Q(i)|}\sum_{q\in Q(i)}\log\frac{\exp(z_i\cdot z_q/\tau)}{\sum_{e\in E(i)}\exp(z_i\cdot z_e/\tau)},
\label{supconloss}
\end{aligned}
\end{equation}
where $E(i) \coloneqq I\textbackslash i$, $Q(i) \coloneqq \{q\in E(i) \mid \widetilde{y}_q=\widetilde{y}_i\}$ represents the set of samples in the batch who has the same label with the $i$-th sample, $\tau \in \mathbb{R}^+$ is a scalar temperature parameter, and $z_i$ is the intermediate representation of $i$ which is extracted from the network.
The overall loss function for the optimization of our model can be formulated as follows:
\begin{equation}
    \mathcal{L} = \mathcal{L}_{\mathrm{cls}} + \lambda\mathcal{L}_{\mathrm{con}},
    \label{sumloss}
\end{equation}
where $\lambda$ is a weight parameter to balance between classification and contrastive learning.

\subsection{Post-processing}

After obtaining the series of output probabilities $P=(p_i)_{i=1}^{N}$, we proceed with macro-expression spotting and micro-expression spotting separately, following \cite{leng2022abpn}. We then describe the micro-expression spotting process, and the macro-expression spotting process is similar. We first obtain all possible micro-expression apex frames $U^{\mathrm{mi},\mathrm{apex}}$ with the rule $p_l^{\mathrm{mi}, \mathrm{apex}} > \theta_{\mathrm{apex}}$, where $\theta_{\mathrm{apex}}$ is a threshold. For each possible apex frame $u^{\mathrm{mi}, \mathrm{apex}}_l\in U^{\mathrm{mi},\mathrm{apex}}$, we select the onset frame with the highest onset probability from the left side of the apex frame within the range of $[l-\frac{k^{\mathrm{mi}}}{2}, l-\frac{j^{\mathrm{mi}}}{2}]$, and we select the offset frame with the highest offset probability from the right side of the apex frame within the range of $[l+\frac{j^{\mathrm{mi}}}{2}, l+\frac{k^{\mathrm{mi}}}{2}]$, where $k^{\mathrm{mi}}$ and $j^{\mathrm{mi}}$ represent the average duration and minimum duration of a micro-expression, respectively. As a result, we obtain one micro-expression proposal $\phi_l$, which contains the onset frame, offset frame, and expression type. Subsequently, we assign a score $s_l=p_b^{\mathrm{mi},\mathrm{onset}}\times p_l^{\mathrm{mi},\mathrm{apex}} \times p_d^{\mathrm{mi},\mathrm{offset}}$ to the micro-expression proposal $\phi_l$, where $b$ and $d$ represent the frame indices of the onset frame and offset frame selected by the rule mentioned above, respectively. 

After obtaining all possible expression proposals, we utilize Non-Maximum Suppression to filter out redundant proposals. Specifically, if the overlap rate of two proposals is higher than $\theta_{\mathrm{overlap}}$, we compare the assigned scores and discard the proposal with the lower score, thereby obtaining the final spotting results.

\begin{table*}[htbp]
\setlength\tabcolsep{4pt}
  \centering
  \caption{Results of the ablation study on the effectiveness of the proposed modules.}
    \begin{tabular*}{0.75\hsize}{@{}@{\extracolsep{\fill}}llcccccc@{}}
    \hline
    & \multicolumn{3}{c}{SAMM-LV}&\multicolumn{3}{c}{CAS(ME)$^2$}\\
    &MaE&ME&Overall&MaE&ME&Overall\\
    \hline
    \textbf{Baseline}&0.3762&0.2222&0.3392&0.3891&0.1607&0.3561\\
    \textbf{+SLO}&0.4279&0.3455&0.3990&0.3973&0.2270&0.3729\\
    \textbf{+SLO+STGCN}&0.4258&0.3520&0.4017&0.4150&0.1846&0.3808&\\
    \textbf{+SLO+STGCN+FLGP}&0.4364&0.3771&0.4173&0.4095&0.2556&0.3865\\
    \textbf{+SLO+STGCN+FLGP+SupCon}&\textbf{0.4631}&\textbf{0.4035}&\textbf{0.4454}&\textbf{0.4340}&\textbf{0.2637}&\textbf{0.4154}\\
    \hline
    \end{tabular*}%
  \label{results3}%
\end{table*}%

\begin{table}[htbp]
\setlength\tabcolsep{4pt}
  \centering
  \caption{Results of the ablation study on different temporal window sizes (receptive fields).}
    \begin{tabular}{cccc}
    \hline
    Window Size / & \multirow{2}{*}{MaE}&\multirow{2}{*}{ME}&\multirow{2}{*}{F1-Score} \\
      Receptive Field\\
    \hline
    5&0.3627&0.2131&0.3279\\
    9&0.3926&0.3529&0.3818\\
    11&0.4051&0.3467&0.3892\\
    15&0.4337&0.3602&0.4128\\
    \textbf{17}&\textbf{0.4631}&\textbf{0.4035}&\textbf{0.4454}\\
    19&0.4419&0.3873&0.4264\\
    21&0.4202&0.3712&0.4054\\
    25&0.4060&0.3721&0.3955\\
    29&0.3973&0.3610&0.3857\\
    \hline
    \end{tabular}%
  \label{results2}%
\end{table}%

\begin{table}[htbp]
\setlength\tabcolsep{4pt}
  \centering
  \caption{Results of the ablation study on the choice of hyperparameter.}
    \begin{tabular}{cccc}
    \hline
    $\lambda$ & SAMM-LV&CAS(ME)$^2$&Overall \\
    \hline
    0.0&0.4173&0.3865&0.4039\\
    0.005&0.4303&0.3973&0.4160\\
    0.01&0.4337&0.4057&0.4218\\
    0.05&\textbf{0.4454}&0.4154&\textbf{0.4328}\\
    0.08&0.4289&\textbf{0.4167}&0.4238\\
    0.10&0.4194&0.4113&0.4160\\
    \hline
    \end{tabular}%
  \label{resulthyper}%
\end{table}%

\subsection{Training details}
We use the AdamW optimizer \cite{loshchilov2017decoupled} to optimize our model, setting the learning rate to 0.01, $\beta_{1}$ to 0.5, and $\beta_{2}$ to 0.9. The window length $w$ used for partitioning the videos is set to 17, which is equal to the temporal receptive field of our network. The temperature parameter $\tau$ in (\ref{supconloss}) is set to 0.5. We train our model for 100 epochs with a batch size of 512.

\section{EXPERIMENTS}
\subsection{Dataset}
We evaluate our method following the protocol of MEGC2021 on two benchmark datasets: SAMM-LV \cite{yap2020samm} and CAS(ME)$^2$ \cite{qu2017cas}. The SAMM-LV dataset comprises 147 raw long videos containing 343 macro-expression clips and 159 micro-expression clips. The CAS(ME)$^2$ dataset includes 87 raw long videos with 300 macro-expression clips and 57 micro-expression clips. The frame rate of the SAMM-LV dataset is 200fps, while the frame rate of the CAS(ME)$^2$ dataset is 30fps. To align the frame rates of both datasets, we subsample every 7th frame from the SAMM-LV dataset to achieve a frame rate close to 30fps. Leave-one-subject-out cross-validation strategy is utilized in our experiments.

\subsection{Evaluation metric}
We use the evaluation metrics outlined in the MEGC2021 protocol. The true positive (TP) expression proposal in a video is defined based on the intersection between the proposal and the ground-truth expression clip. Specifically, given a ground-truth expression clip and its expression type, we compare it with all expression proposals with the same expression type. An expression proposal $W_{\mathrm{Proposal}}$ is considered TP when it satisfies the following condition:
\begin{equation}
\begin{aligned}
    \frac{W_{\mathrm{Proposal}}\cap W_{\mathrm{GroundTruth}}}{W_{\mathrm{Proposal}}\cup W_{\mathrm{GroundTruth}}}\geq\theta_{\mathrm{IoU}},
\end{aligned}
\end{equation}
where $\theta_{\mathrm{IoU}}$ is set to 0.5 and $W_{\mathrm{GroundTruth}}$ represents the ground-truth expression proposal (from the onset frame to the offset frame). Otherwise, the proposed expression proposal is considered a false-positive (FP). All ground-truth expression clips that do not match any proposal are considered false-negative (FN). According to the protocol, each ground-truth expression clip corresponds to at most one TP. We calculate the precision rate, recall rate, and F1 score to evaluate the performance of our model.

\subsection{Ablation studies}
\textbf{Effectiveness of proposed modules.} We first validate our proposed modules and Table \ref{results3} shows the experimental results. We report the F1-score for macro-expression (MaE) spotting, micro-expression (ME) spotting, and overall performance. \textbf{Baseline} in Table \ref{results3} refers to the method proposed in \cite{leng2022abpn}. The acronyms \textbf{SLO}, \textbf{STGCN}, \textbf{FLGP}, and \textbf{SupCon} correspond to specific modifications in our method. Specifically, \textbf{SLO} represents computing short- and long-term optical flows with our receptive field adaptive sliding window strategy instead of computing optical flows between adjacent frames, \textbf{STGCN} involves employing spatio-temporal GCNs for feature learning instead of flattening the graph-structured data and learning the features using CNNs, \textbf{FLGP} introduces facial local graph pooling into our ST-GCN network for multi-scale feature learning, and \textbf{SupCon} introduces supervised contrastive loss for discriminative feature learning.
The results demonstrate the effectiveness of our proposed modules. Notably, the receptive field adaptive sliding window strategy enhances the extraction of motion features, particularly in magnifying subtle motions that exist in micro-expressions. This enhancement leads to a significant overall performance improvement of 17.6\%/4.7\% on the SAMM-LV and CAS(ME)$^2$ datasets.
The utilization of ST-GCN and its combination with facial local graph pooling for multi-scale feature learning results in a further improvement of 4.6\%/3.6\% on the SAMM-LV and CAS(ME)$^2$ datasets compared to using one-dimensional CNNs. This demonstrates the superior representational capabilities of our proposed model.
Finally, the introduction of the supervised contrastive loss enables our model to better recognize the boundaries of distinguishing between different types of expressions, resulting in an improvement of 6.7\%/7.5\% on the SAMM-LV and CAS(ME)$^2$ datasets.

\begin{figure*}[tbp]
\centering
\includegraphics[width=0.95\textwidth]{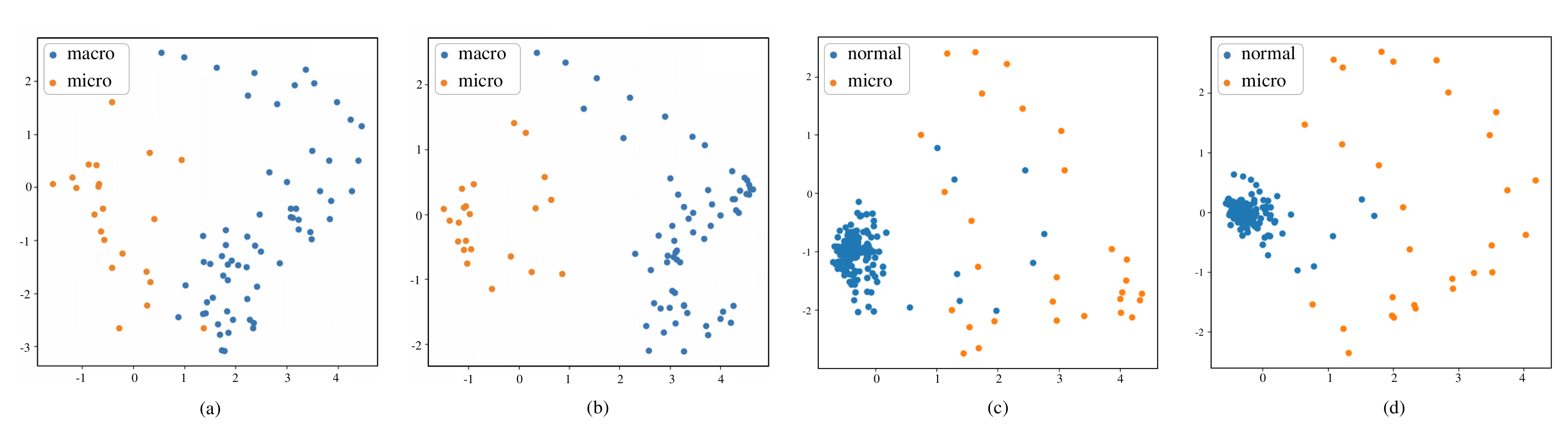}
\caption{Visualization analysis of supervised contrastive learning.
(a) and (b) show the PCA distribution of certain macro- and micro-expression frames: (a) without supervised contrastive learning and (b) with supervised contrastive learning.
(c) and (d) depict the PCA distribution of certain micro-expression frames and normal frames: (c) without supervised contrastive learning and (d) with supervised contrastive learning.}
\label{fig6}
\end{figure*}
\begin{figure*}[tbp]
\centering
\includegraphics[width=0.95\textwidth]{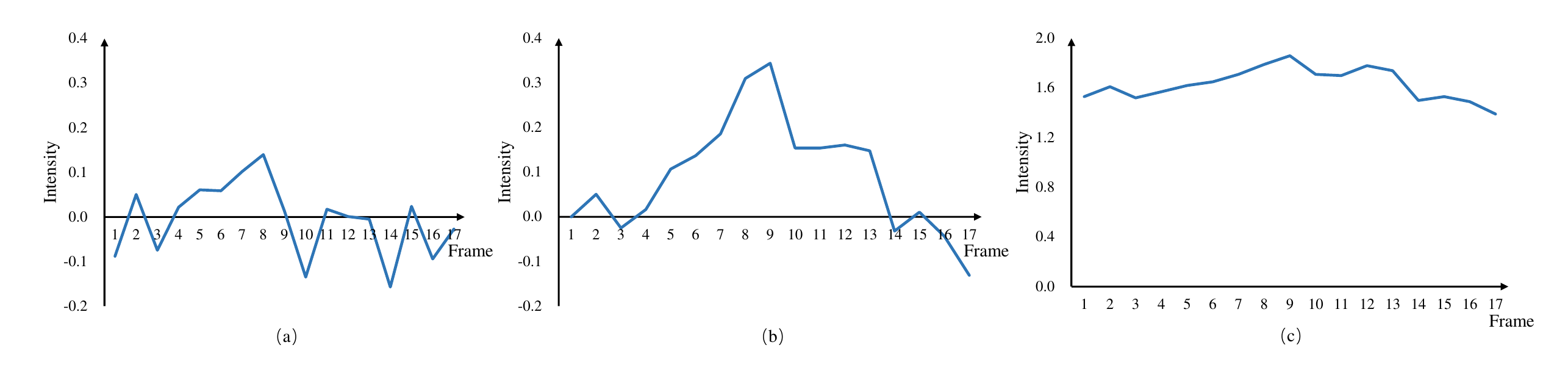}
\caption{Some visualization optical flows of certain micro-expression frames computed by three strategies. The data comes from the vertical component of optical flows computed at the left mouth corner when subject 11 from the SAMM-LV dataset is performing a micro-expression. (a) optical flows computed between adjacent frames; (b) optical flows computed with our receptive field adaptive sliding window strategy; (c) optical flows computed with a large sliding window strategy.}
\label{fig7}
\end{figure*}

In addition to the quantitative results, Fig.~\ref{fig6} and Fig.~\ref{fig7} show some qualitative analyses of our proposed receptive field adaptive sliding window strategy and supervised contrastive learning. 
Fig.~\ref{fig6} illustrates the visualization analysis of introducing supervised contrastive learning into our model. During inference, we randomly sampled some frames and employed principal component analysis (PCA) \cite{bro2014principal} to examine the distribution of various expression classes. Subsequently, we labeled each frame with its ground-truth expression label. In Fig.~\ref{fig6} (a) and (b), we compare the PCA distribution of specific macro-expression frames and micro-expression frames with and without the use of supervised contrastive learning. In Fig. \ref{fig6}~(c) and (d), we compare the PCA distribution of specific normal frames and micro-expression frames with and without the use of supervised contrastive learning. The results indicate that when we do not introduce supervised contrastive learning into our model, the distribution of certain frames from different classes might become mixed, resulting in misclassification. However, after introducing supervised contrastive learning to our model, the domain discrepancy increases, leading to improved accuracy in expression spotting.
Fig.~\ref{fig7} presents visualizations of optical flows calculated using three different strategies. When computing optical flows between adjacent frames, the motions that exist in micro-expressions are so subtle that the noise in the optical flows might overshadow these delicate movements, impacting the quality of the data and making it difficult to reveal these subtle expressions.
On the other hand, using a large sliding window strategy to compute optical flows can introduce significant influence from head movements, making it difficult to analyze expressions accurately. However, the utilization of our receptive field adaptive sliding window strategy helps alleviate these problems. It strikes a balance between magnifying subtle expression motions and minimizing the influence of head movements.

\begin{table*}[htbp]
\setlength\tabcolsep{4pt}
  \centering
  \caption{Comparison with the state-of-the-art methods on CAS(ME)$^2$ and SAMM-LV in terms of F1-score.}
    \begin{tabular*}{0.75\hsize}{@{}@{\extracolsep{\fill}}llcccccc@{}}
    \hline
    \multicolumn{2}{c}{Methods} & \multicolumn{3}{c}{SAMM-LV} & \multicolumn{3}{c}{CAS(ME)$^2$} \\
    &&MaE&ME&Overall&MaE&ME&Overall \\
    \hline
    Traditional methods&MDMD\cite{he2020spotting}&0.0629 &0.0364&0.0445&0.1196&0.0082&0.0376\\
    &Optical Strain\cite{gan2020optical}&-&-&-&0.1436&0.0098&0.0448\\
    &Zhang et al.\cite{zhang2020spatio}&0.0725&0.1331&0.0999&0.2131&0.0547&0.1403\\
    &He \cite{yuhong2021research}&0.4149&0.2162&0.3638&0.3782&0.1965&0.3436\\
    &Zhao et al.\cite{zhao2022rethinking}&-&-&0.3863&-&-&0.4030\\
    \hline
    Deep-learning methods&Verburg\cite{verburg2019micro}&-&00821&-&-&-&-\\
    &LBCNN\cite{pan2020local}&-&-&0.0813&-&-&0.0595\\
    &MESNet\cite{wang2021mesnet}&-&0.0880&-&-&0.0360&-\\
    &SOFTNet\cite{liong2021shallow}&0.2169&0.1520&0.1881&0.2410&0.1173&0.2022\\
    &3D-CNN\cite{yap20223d}&0.1595&0.0466&0.1084&0.2145&0.0714&0.1675\\
    &Concat-CNN\cite{yang2021facial}&0.3553&0.1155&0.2736&0.2505&0.0153&0.2019\\
    &LSSNet\cite{yu2021lssnet}&0.2810&0.1310&0.2380&0.3770&0.0420&0.3250\\
    &MTSN\cite{liong2022mtsn}&0.3459&0.0878&0.2867&0.4104&0.0808&0.3620\\
    &ABPN\cite{leng2022abpn}&0.3349&0.1689&0.2908&0.3357&0.1590&0.3117\\
    &AUW-GCN\cite{yin2023aware}&0.4293&0.1984&0.3728&0.4235&0.1538&0.3834\\
    &\textbf{Ours}&\textbf{0.4631}&\textbf{0.4035}&\textbf{0.4454}&\textbf{0.4340}&\textbf{0.2637}&\textbf{0.4154}\\
    \hline
    \end{tabular*}%
  \label{results1}%
\end{table*}%

\begin{table*}[htbp]
\setlength\tabcolsep{4pt}
  \centering
  \caption{Detailed spotting results of the proposed method on CAS(ME)$^2$ and SAMM-LV.}
    \begin{tabular*}{0.75\hsize}{@{}@{\extracolsep{\fill}}llcccccc@{}}
    \hline
    Dataset & \multicolumn{3}{c}{SAMM-LV}&\multicolumn{3}{c}{CAS(ME)$^2$} \\
      Expression & MaE&ME&Overall&MaE&ME&Overall\\
    \hline
    Total&343&159&502&300&57&357\\
    TP&188&69&257&161&12&173\\
    FP&281&114&395&281&22&303\\
    FN&155&90&245&139&45&184\\
    Precision&0.4009&0.3770&0.3942&0.3643&0.3529&0.3634\\
    Recall&0.5481&0.4340&0.5120&0.5367&0.2105&0.4678\\
    F1-Score&0.4631&0.4035&0.4454&0.4340&0.2637&0.4154\\
    \hline
    \end{tabular*}%
  \label{results4}%
\end{table*}%

\textbf{Temporal window size.} We further explore how much temporal information we need to predict one frame. In practice, we test different temporal window sizes (i.e., the receptive field of our network) on the SAMM-LV dataset, and the experimental results are shown in Table \ref{results2}. The results show that when it comes to 17, the performance achieves the best. This is because the temporal boundary for distinguishing macro- and micro-expressions is 0.5 seconds, which corresponds to 15 frames when the frame rate is set to 30fps. When the receptive field comes to 17, the temporal window is enough to perceive a complete micro-expression and magnify the motion information that exists in micro-expressions. Therefore, it enables the model capture the temporal variations needed to distinguish between general macro-expressions and micro-expressions. When the receptive field becomes larger, the increasing number of frames may lead to information redundancy, and increasingly serious head movement problems may also impact the accuracy of expression spotting.

\textbf{Hyperparameter $\lambda$.} The weight parameter $\lambda$ in (\ref{sumloss}) is set to balance classification and contrastive learning. We conducted experiments with various $\lambda$ values on the SAMM-LV and CAS(ME)$^2$ datasets, and the corresponding results are presented in Table \ref{resulthyper}.
We observed that the optimal value for $\lambda$ varies across different datasets. However, the best overall performance is achieved when $\lambda$ is set to 0.05. Increasing $\lambda$ beyond this value starts to impact standard classification, leading to a decrease in spotting accuracy. 

\subsection{Comparison with state-of-the-art methods}

We compare our method with the state-of-the-art methods on the SAMM-LV and CAS(ME)$^2$ datasets and the results are shown in Table \ref{results1}. For the overall performance, our method achieves F1-scores of 0.4454 on the SAMM-LV dataset and 0.4154 on the CAS(ME)$^2$ dataset, which outperforms other state-of-the-art methods by 15.30\% and 3.08\% respectively. For the macro-expression spotting, our method achieves an improvement of 7.8\%/2.5\% on the SAMM-LV and CAS(ME)$^2$ datasets respectively compared to other methods.
It is important to emphasize our method's remarkable effectiveness in micro-expression spotting. The results demonstrate a substantial enhancement with an 86.63\% improvement on the SAMM-LV dataset and a 34.20\% improvement on the CAS(ME)$^2$ dataset in comparison to other state-of-the-art methods. The results show our method's ability to capture subtle motions that exist in micro-expressions and alleviate the impact of irrelevant motions.

\subsection{Detailed discussion}
\label{secd}
Table \ref{results4} shows the detailed results on the SAMM-LV and CAS(ME)$^2$ datasets. It can be seen that our method achieves a recall rate of approximately 0.5 in MaE spotting and ME spotting on the SAMM-LV dataset, as well as in MaE spotting on the CAS(ME)$^2$ dataset. 
The particular issue in not achieving a higher recall rate for ME spotting on the CAS(ME)$^2$ dataset may be due to several factors. One primary factor could be the limited amount of data available, as the CAS(ME)$^2$ dataset contains only 57 ME clips. Additionally, some of these clips may involve eye blinking, which we consider as irrelevant motion. Addressing these issues, including the consideration of generating more diverse micro-expression data, will be a focal point of our future research.

\section{CONCLUSIONS AND FUTURE WORKS}


In this paper, we presented a novel framework for macro- and micro-expression spotting. We proposed a receptive field adaptive sliding window strategy to compute short- and long-term optical flow to magnify the motion information existed in facial expressions. Furthermore, we proposed SpoT-GCN to fully capture the spatial and temporal relationships that existed in the optical flow features, where the introduction of FLGP enables the network to learn multi-scale features. In addition, we introduced supervised contrastive loss for more discriminative feature representation learning. Comprehensive experiments were conducted on the SAMM-LV and CAS(ME)$^2$ datasets to verify the effectiveness of our proposed method.


In the future, as mentioned in Section \ref{secd}, we hope to introduce a method for generating diverse MEs in order to enrich the ME dataset. Moreover, we aim to develop a deep learning-driven technique for motion extraction and create an end-to-end framework for facial expression spotting.





{\small
\bibliographystyle{ieee}
\bibliography{main}
}

\end{document}